\documentclass[10pt,twocolumn,letterpaper]{article}

\usepackage[usenames,dvipsnames,svgnames,table]{xcolor}
\usepackage{cvpr}
\usepackage{times}
\usepackage{epsfig}
\usepackage{graphicx}
\usepackage{amsmath}
\usepackage{amssymb}
\usepackage{mathrsfs}
\usepackage[numbers]{natbib}
\usepackage{booktabs}
\usepackage{soul}
\usepackage{xspace}
\usepackage{algorithm}
\usepackage[noend]{algpseudocode}
\usepackage{subcaption}
\usepackage{multirow}
\usepackage{tabularx}

\newcommand{\comment}[1]{}

\newcommand{\amosrmk}[1]{}
\newcommand{\amos}[1]{#1}

\newcommand{\jacquesrmk}[1]{}

\newcommand{\nicolasrmk}[1]{}

\newcommand{\davidermk}[1]{}

\newcommand{\vincentrmk}[1]{}
\newcommand{\stephane}[1]{#1}
\newcommand{\stephanermk}[1]{}
\newcommand{\Eb}{Event-based\ }
\newcommand{\eb}{event-based\ }

\newcommand{\e}{\mathbf{e}}

\newcommand{\s}{\mathbf{s}}

\newcommand{\N}{\mathbb{N}}

\newcommand{\F}{\mathcal{F}}
\newcommand{\D}{\mathbf{D}}

\newcommand{\evHarris}{{\it evHarris\ }}
\newcommand{\evFast}{{\it evFast\ }}
\newcommand{\Arc}{{\it Arc\ }}
 
\cvprfinalcopy 
\def\cvprPaperID{4920} 

\ifcvprfinal\fi
\begin{document}

\title{Speed Invariant Time Surface\\ for Learning to Detect Corner
  Points with  Event-Based Cameras}

\author{
	Jacques Manderscheid$^1$, Amos Sironi$^1$, Nicolas Bourdis$^1$, Davide Migliore$^1$ and Vincent Lepetit$^2$\\
	$^1$Prophesee, Paris, France, $^2$University of Bordeaux, Bordeaux, France\\
	{\tt\small \{jmanderscheid, asironi, nbourdis, dmigliore\}@prophesee.ai, vincent.lepetit@u-bordeaux.fr}
					}

\maketitle
\vspace{-3cm}

\begin{abstract}
			
	We propose a learning approach to  corner detection for \eb cameras that
        is stable  even under fast and  abrupt motions.  \Eb cameras  offer high
        temporal resolution, power efficiency, and high dynamic range.  However,
        the  properties of  \eb data  are  very different  compared to  standard
        intensity  images, and  simple  extensions of  corner detection  methods
        designed for  these images do  not perform well  on \eb data.   We first
        introduce an efficient  way to compute a time surface  that is invariant
        to the speed  of the objects.  We  then show that we can  train a Random
        Forest to  recognize events generated by  a moving corner from  our time
        surface.  Random Forests  are also extremely efficient,  and therefore a
        good choice to deal with the high capture frequency of \eb cameras---our
        implementation  processes up  to 1.6Mev/s  on a  single CPU.
        Thanks to our  time surface formulation and this  learning approach, our
        method is  significantly more robust  to abrupt changes of  direction of
        the  corners  compared to  previous  ones.   Our method  also  naturally
        assigns a  confidence score  for the  corners, which  can be  useful for
        postprocessing.   Moreover,  we   introduce  a  high-resolution  dataset
        suitable for quantitative evaluation  and comparison of corner detection
        methods for \eb cameras.  We call our approach SILC, for Speed Invariant
        Learned Corners, and  compare it to the  state-of-the-art with extensive
        experiments, showing better performance.

\end{abstract}

\vspace{-5mm}

\section{Introduction}
\label{sec:introduction}

\begin{figure}
	\centering
	\includegraphics[width=1\linewidth,height=0.8\linewidth]{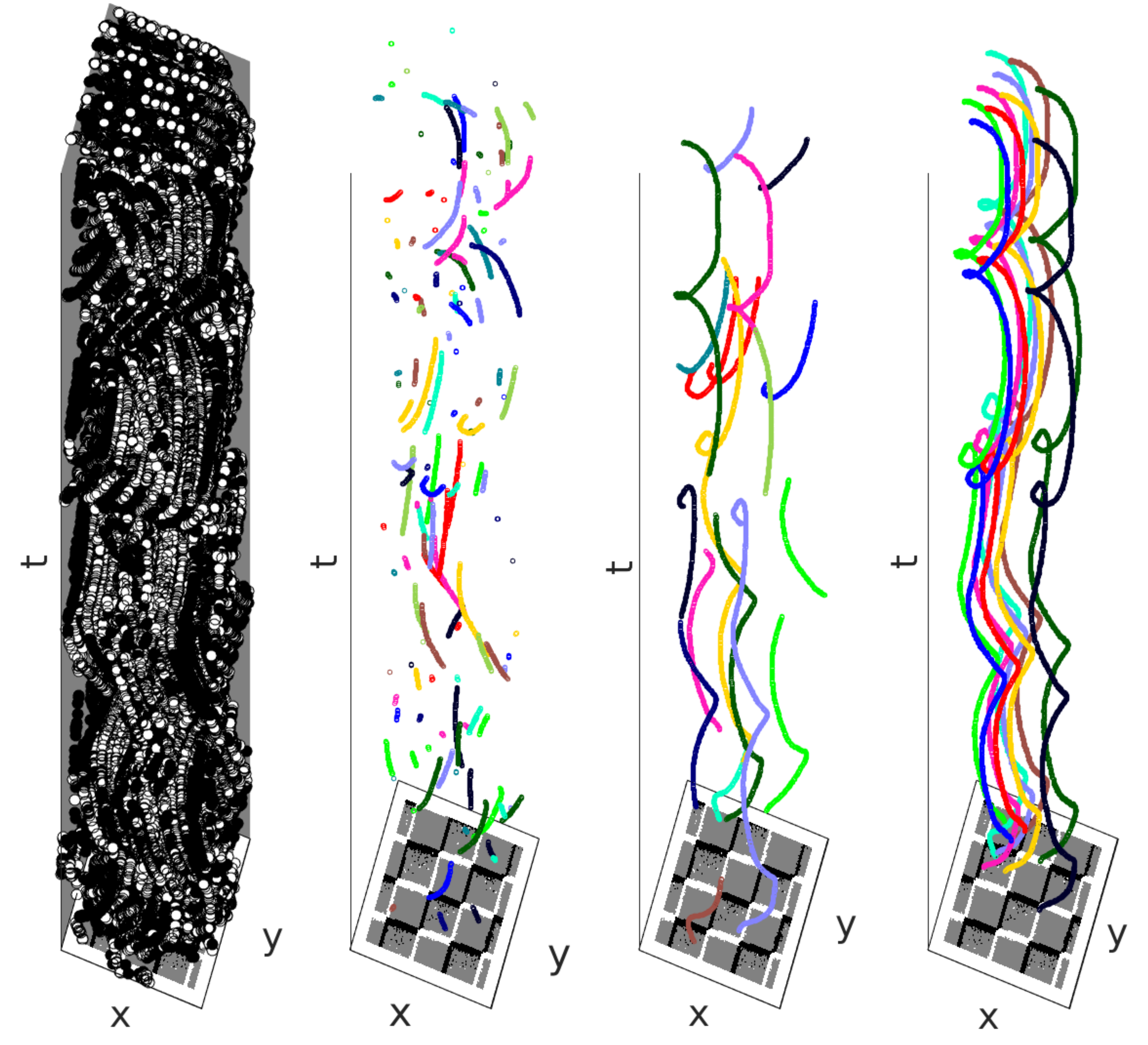}
	\begin{tabular}{cccc}
		\hspace{-0.4cm}  {\small { \bf (a) Events   }       }       &   
		\hspace{-0.15cm}{\small {\bf (b) evFAST~\cite{Mueggler17} }}&   
		\hspace{-0.2cm}{\small{\bf (c)  evHarris~\cite{Vasco16}    }}&   
		\hspace{-0.1cm}{\small{\bf (d) Ours }}		        					\end{tabular}
		\caption{{\bf(a)} Stream of events generated by  an \eb camera moving in front
		of a  checkerboard pattern.   Black dots represents  events with  a negative
		polarity, white dots events with  a positive polarity. {\bf(b-c)} Standard \eb
		corner detectors~\cite{Mueggler17,Vasco16}  are not  robust to direction  changes of
		the camera, and  the corners cannot be reliably tracked  over time without a
		very complex tracking  scheme.  {\bf(d)} By training a  classifier to detect
		corners from  \eb data, our  method can  reliably detect corners  under even
		abrupt  changes of  direction. A  simple nearest  neighbor tracker  produces
	continuous trajectories of the corners over time.}
	\vspace{-3mm}
	\label{fig:intro}
\end{figure}

By  capturing   very  efficiently  local  illuminance   changes~('events'),  \eb
cameras~\cite{Lichtsteiner08,Posch11,Serrano13} open the door to novel very fast
and  low-power  computer vision  algorithms  able  to  deal with  large  dynamic
ranges~\cite{Benosman14, Kim16, Rebecq17,Andreopoulos18}.   However, because the
events are created asynchronously,  as shown in Fig.~\ref{fig:intro}~{(a)}, novel
algorithms have  to be  developed to perform  fundamental computer  vision tasks
that are typically performed on regular frame images.

One of the main fundamental tasks is feature point detection, which is important
for applications with very strong dynamics  such as UAV navigation, where motion
blur makes classical  frame-based approaches less robust, or  visual odometry in
High  Dynamic  Range~(HDR)  conditions,  among others.   Inspired  by  the  vast
literature  on   frame-based  feature   point  detection,  some   works  adapted
frame-based corner detector to \eb  data.  Typically, a local spatial descriptor
is built  around an  event, for  example by  cumulating events  in a  given time
window~\cite{Vasco16},  or   by  considering  the   times  of  arrival   of  the
events~\cite{Mueggler17}.  Then,  a classical test, such  as~\cite{Harris88} can
be applied to this 2D spatial neighborhood.  However, the resulting detectors do
not take  into consideration the specific  characteristics of \eb data,  such as
different  noise  patterns,  responses  to changes  of  direction,  illumination
changes, etc.  Even  if efforts have been  made in order to  design better tests
for  \eb  cameras~\cite{Mueggler17,Alzugaray18}, hand-crafted  detectors  remain
unstable   and  corners  can  not   be   reliably   detected   over
time Fig.~\ref{fig:intro}~{(b-c)}.

In this  paper, we  propose a  learning approach to  \eb feature  detection.  We
train a classifier  to label individual events as generated  by a moving feature
point or  not.  The main  advantage of taking a  learning approach is  to obtain
more stable corners:  As shown in Fig.~\ref{fig:intro}, a typical  error made by
previous detectors is  that  they are  sensitive to  changes of  the
apparent motion of  the feature points.  This is because  corners in \eb cameras
are  not  invariant under  changes  of  direction,  by  contrast to  corners  in
intensity images.  Previous detectors also often erroneously detect points along
edges because of noisy events, while such  points cannot be detected in a stable
way.  Learning  makes the  detection more  robust to  motion changes  and noise,
without having to manually design an \emph{ad hoc} method.

Our  classification  approach  relies  on   a  novel  formulation  of  the  Time
Surface~\cite{Benosman12,Mueggler17},  which  is  another contribution  of  this
work.  The Time Surface is a representation that accumulates the information from events
over time, and  is a common tool used in \eb vision, including to detect
corner  points.  In  our work,  we also  use  a Time  Surface as  input to  the
classifier,  but we  show  how to  efficiently  create a  Time  Surface that  is
invariant  to the  objects'  speed.   Previous work~\cite{Alzugaray18b}  already
introduced a method for computing a  time surface invariant to speed, however
it is still too slow  to  compute, 
and  incompatible  with the  high  frequency of  \eb
cameras.  The  invariance to speed of  our Time Surface is  important both to achieve
classification  performance and also to  keep  the classifier  small, 
which makes computation fast.

One critical aspect of our learning-based approach is indeed that classification
must be performed  extremely fast, otherwise we would lose  the advantage of the
capture efficiency of  \eb cameras.  We therefore chose to  use a Random Forest,
as Random Forests  are very efficient without  having to use a  GPU (unlike Deep
Networks),  which   would  be  counter-productive  since   we  target  low-power
applications. In fact, parallelizing computation as done by  GPUs is not well adapted
to  the  sparse  and  asynchronous  nature of  the  events. 
Our  current implementation processes up to $1.6\cdot10^6$ events
per second on  a single CPU.

To evaluate the  quality of our detector, we also  release a new high-resolution
benchmark dataset.   We propose a  metric which  is independent of  ground truth
keypoints  extracted from  gray level  images, which  are often  used but  would
introduce  a  strong bias  in  the  evaluation.   Specifically, we  compare  our
approach to  different detectors in  combination with a simple  nearest neighbor
based tracking, showing that, thanks to the temporal continuity of the events, a
very simple tracking rule can lead to state-of-the-art results.

\vspace{-1mm}
\section{Event-based cameras}
\label{sec:event-based_cameras}
\vspace{-1mm}
Standard frame-based  cameras capture visual information  by acquiring snapshots
of the observed scene at a fixed rate. This might result 
in motion blur for highly dynamic scenes
and in redundant data generation for static ones.
By contrast,  \eb cameras~\cite{Lichtsteiner08,Posch11,Serrano13},  a relatively
recent type of cameras, adaptively record information from a scene, depending on
the content.   More precisely, each pixel  in an \eb sensor  is independent from
the rest of the pixel array.  When a change in the log-luminosity intensity at a
location surpasses  a threshold, the  corresponding pixel emits  an \textit{event}.
An event contains  the 2D location of  the pixel, the timestamp  of the observed
change and its polarity, \ie a binary variable indicating whether the luminosity
increased or decreased.
Thanks  to  these characteristics,  \eb  cameras  have temporal
resolution in the order of the microsecond, high dynamic  range,
and are  well suited  for low-power
applications.

Some types of \eb cameras also provide gray-level measurements together with the
change detection events.  The DAVIS~\cite{Brandli14} sensor for example, is a $240\times180$
sensor able
to   output   intensity   frames   at    standard   frame   rates.    The   ATIS
sensor~\cite{Posch11}  instead provides  asynchronous intensity  measurements in
form of time differences at the same temporal resolution of the events.
Our  method can  be applied  to any  type of  \eb cameras  and does  not require
gray-level information at run time.  Intensity information has been used in this
work only  to build the  dataset used to train  our classifier, as  explained in
Section~\ref{subsec:train_dataset}.

Finally, we note that the spatial resolution of
\eb cameras, still limited to low resolution, is 
likely to soon reach standard resolutions of frame-based
cameras~\cite{Son17}.  As a consequence, \eb algorithms will have to 
process constantly increasing event rates.
In this  work, we  use and  release the first  \eb dataset  acquired with  a HVGA
(480$\times$360) sensor, showing  that our algorithm is suited to  real time and
high data rate applications.
 
\section{Related work}
\label{sec:related_work}
\vspace{-1mm}
In this section, we review related work on \eb feature detection,
machine learning approaches applied to \eb cameras and to feature point detection.
\vspace{-3mm}
\paragraph{\Eb Features Detection}
\Eb feature  detection methods can  be divided  into two categories.   The first
class of  methods~\cite{Clady15,Ni15,Lagorce14} aims  at detecting  a particular
pattern in the stream of events  by applying local template matching rules.  For
example, a  blob detector~\cite{Lagorce14} can  be obtained by using  a gaussian
correlation kernel.  When  enough events are received in the  receptive field of
the kernel, a  feature event is generated. The  method is
generalized to generic  shapes, like corners or T-junctions. 
In~\cite{Clady15}  instead, corners are identified  as intersection of
local  velocity  planes,   which  are  obtained  by  fitting  a   plane  on  the
spatio-temporal  distribution  of  events.   This   class  of  methods  is  very
efficient, but it is sensitive to noise and requires careful tuning of the model
parameters.
		
\begin{figure}
	\centering
	\includegraphics[width=0.9\columnwidth,height=0.53\columnwidth]{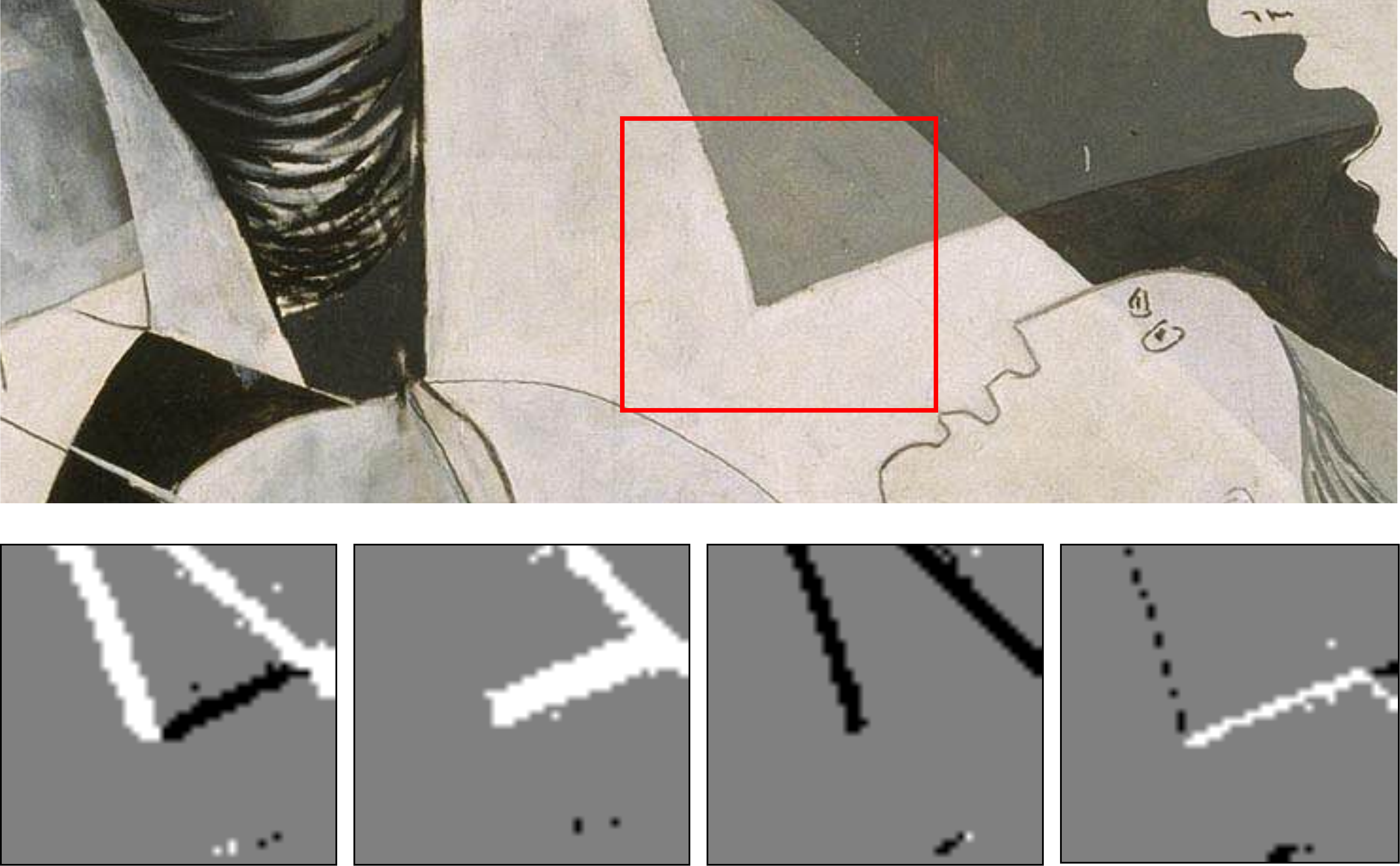}
	\caption{{\bf (Top)} In a classical frame-based camera, the appearance of a corner, 
		such as the one in the red square, is invariant under camera motions.
		{\bf (Bottom)} In the case of an \eb camera, instead, the same corner
		can generate vert different pattern of the events depending on its motion.
		The four panels show 40ms of events generated by the corner on the top
	while rotating the camera at different speeds.}
	\vspace{-3mm}
	\label{fig:corner_variability}
\end{figure}

The  second  class of  methods  instead,  relies  on the  adaptation  of
classical frame-based corner detectors to \eb data.  The basic principle
behind these  approaches is  to build a  local representation  around an
event  and  then apply  classical  corner  tests  on it.   For  example,
in~\cite{Vasco16}  the  Harris  corner  detector is  applied  to  images
obtained  by cumulating  a fixed  number of  events in  a local  spatial
window.   The  Surface of  Active  Events  (or Time  Surface,  presented in
Sec.~\ref{subsec:time_surface})  is a  common representation  used in
\eb  vision~\cite{Benosman12,Mueggler17,Zhu18,Sironi18}.   It  has  been
used in~\cite{Mueggler17}  for corner  detection, where the  FAST 
algorithm~\cite{Rosten06} is adapted to the  pattern of Time Surface of a
moving corner.  In~\cite{Alzugaray18} this formulation has been extended
to corners with obtuse angles.
These methods  lose accuracy mostly in  case of changes of  feature point motion
direction, because  \eb data do not  share the same characteristics  as standard
intensity images.   In particular, the corner  appearance in a stream  of events
depends  on its  motion, as  shown in  Fig.~\ref{fig:corner_variability}.  As  a
consequence,  the constant  appearance  assumption made  in frame-based vision
does not hold in the event-based case.
		
To   overcome   this   problem,   some  works   rely   on   gray   level
images~\cite{Tedaldi16,Kueng16,Liu16,Gehrig18}.   In  these  approaches,
feature points are detected in the images using standard techniques, and
then tracked in between by  using the event stream.  These methods
have the same  limitation as frame-based approaches,  namely they lose
accuracy in presence of motion  blur, HDR situations, and increase data
rate and the power consumption.   Moreover, they require either an \eb
sensors  able   to  acquire   graylevel  information,  or   a  careful
synchronization  of  \eb and  frame-based  sensors.   By contrast, our
method requires at run time only  a single and purely \eb sensor.   We claim that
the time  information carried  by the events  is sufficient  to reliably
detect feature  points, without the  need of intensity  information.  We
propose to  use a machine learning  approach to learn the  appearance of
the features directly from the input data.

Finally,    many    works   focus    on    the    problem   of    \eb    feature
tracking~\cite{Drazen11,Ni12,Kueng16,Glover17,Zhu17,Alzugaray18b}.   Our  method
can be combined  with any of these tracking algorithms.   However, thanks to the
high temporal  resolution of  the \eb  cameras, given  our stable  and efficient
feature detector,  the problem of  data association is  made much simpler  and a
simple nearest neighbor matching rule is sufficient in our experiments to obtain
accurate results. 

 \vspace{-3mm}
\paragraph{Learning from Events}
Machine  learning  approaches  for  \eb  cameras can  also  be  divided  in  two
categories.                    In                    the                   first
category~\cite{Li16,Peng17,Clady17,Sironi18,Liu18,Maqueda18,Zhu18,Cannici18},
the events from the camera are accumulated for  a given period of time, or for a
fixed number  of events, to  build a dense representation.   This representation
can then be  effectively processed by standard Computer  Vision techniques, like
Convolutional Neural Networks~(CNNs).
		
In~\cite{Maqueda18} for example, the input events are summed into histograms and
the obtained images are used to predict the steering angle of a car using a CNN.
Similarly, in~\cite{Zhu18}, histograms and time surfaces are used to predict the
optical flow.   In~\cite{Clady17,Ramesh17,Sironi18},  dense
descriptors are  built by comparing and  averaging the timings of  the events in
local windows. The descriptors  are then passed to a classifier,  such as an SVM
or a  neural network,  for object  recognition or other  vision tasks.   Even if
these approaches  reach good  performance and are  efficient, they  increase the
latency of the system by requiring an artificial accumulation time.
		
A  second class  of  methods  avoids this  limitation  by  processing the  input
event-by-event~\cite{Gutig06,Linares11,Oconnor13,Neil16,Lagorce17}.    The  most
common     model    used     in     \eb    cameras     are    Spiking     Neural
Networks~\cite{Oconnor13,Masquelier07,Bichler12,Sheik13,Marti16,Bohte02,Russell10,Cao15},
which have been  proven to reach good accuracy for  simple classification tasks.
However, these approaches are difficult to scale because of the large event rate
of    the   \eb    cameras.    Hardware    implementations   and    neuromorphic
architectures~\cite{Caviar09,Furber14,Akopyan15,Davies18} have  been proposed to
overcome this  bottleneck. However, these  architecture are not large  enough to
support large networks and high-resolution \eb cameras as input.
		
By contrast our classifier is  applied event-by-event, keeping the original time
resolution of the camera, while running on a standard CPU.
	  
\vspace{-2mm}
\paragraph{Learning Frame-based Keypoint Detection.}
A  number of  works  in frame-based  computer  vision focus  on  the problem  of
learning feature  points, as  we do here  for \eb computer  vision.  One  of the
first   approaches   using   machine    learning   for   corner   detection   is
FAST~\cite{Rosten06}, where  a decision  tree is  used to  generalize brightness
tests  on  contiguous  circular  segments  introduced  in~\cite{Rosten05}.   The
motivation for  using learning in  FAST was to speed  up the detection  by early
rejection of non-corner points.  Other authors~\cite{Sochman09,Di18} also showed
how  to learn  a fast  approximation of  existing detectors.   But other  authors
follow this  direction by training  efficient ensembles of decision  trees.  For
example, in~\cite{Strecha09}, the WaldBoost  algorithm~\cite{Sochman05}
is used to learn detectors for specific tasks.
	
More recent works rely on Deep Learning models~\cite{Verdie15,Altwaijry16,Yi16}.
For example, in~\cite{Verdie15}, a regressor  is trained to detect stable points
under  large illumination  changes.   In~\cite{Detone18}, a  fully-convolutional
model is  trained with self-supervision  to jointly predict interest  points and
descriptors.

\section{Method}
\label{sec:method}

\begin{figure*}[tp]
	\centering
				\includegraphics[width=0.95\linewidth,height=0.3\linewidth]{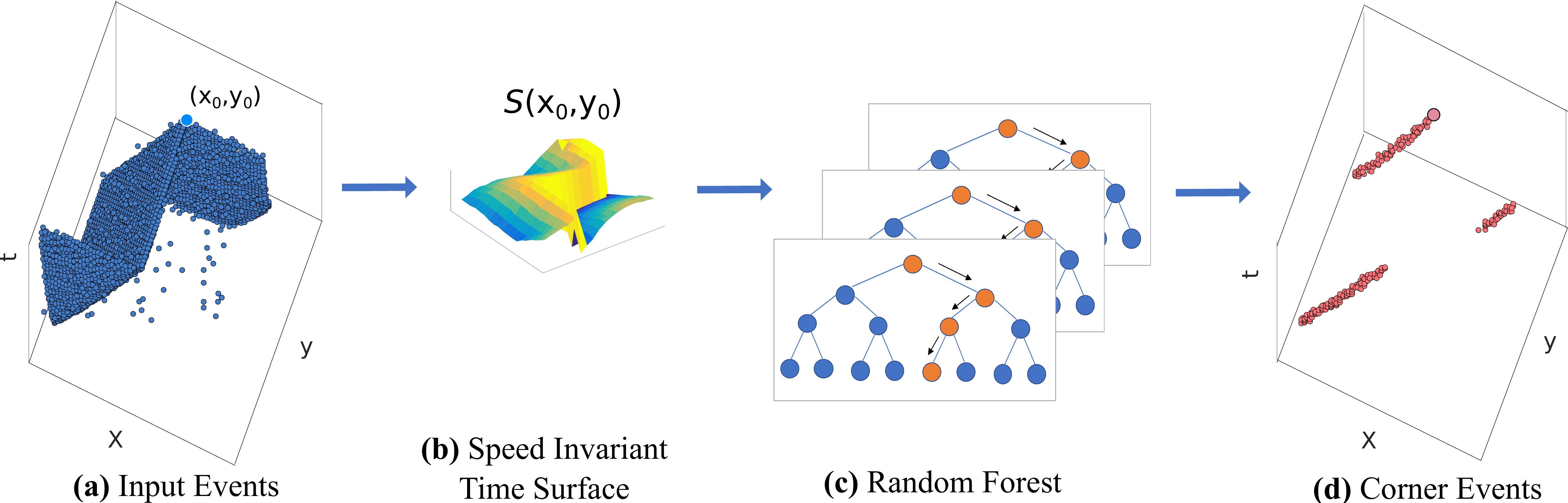}			\caption{Overview of the proposed method. For each incoming event $(x_0,y_0,t_0,p_0)$
		in the input stream {\bf (a)} we compute the Speed Invariant Time Surface {\bf (b)}.
		The Speed Invariant Time Surface is used as input to a Random Forest 
		trained to discriminate corner points {\bf (c)}.
		If the probability returned by the Random Forest is above a threshold, the event is
		classified as a corner. 
		This generates a sparse and stable stream of corner events {\bf (d)}
		which can be used for further processing.	}
	\vspace{-4mm}
	\label{fig:method_overview}
\end{figure*}

In this section, we first introduce  our speed invariant formulation of the time
surface, then we  formalize the problem of learning a  feature detector from the
output  of an  \eb camera.  Finally,  we explain  how  to create  a dataset  for
training  a corner  events detector.   An  overview of  our method  is given  in
Fig.~\ref{fig:method_overview}.

\subsection{Asynchronous Event-based Representation of a Visual Scene}
\label{subsec:event_representation}

As  described in  Section~\ref{sec:event-based_cameras},  the output  of an  \eb
camera  is an  asynchronous stream  of events  $\{\e_i\}_{i\in\N}$.  Each  event
$\e_i$ represents a contrast change at a given pixel and at a given time and can be
formalized as
\begin{equation}
	\e_i = (x_i, y_i, t_i, p_i) \> ,
	\label{eq:event}
\end{equation}
where $(x_i,y_i)$  are the  pixel coordinates  of the  event, $t_i\geq0$  is the
timestamp at which the event was generated, and $p_i\in\{-1,1\}$ is the polarity
of  the event,  with  $-1$ and  $+1$ meaning  respectively  that the  luminosity
decreased or increased at that pixel.

Given an input event $\e_i$, \eb corner detectors typically rely on
a spatio-temporal neighborhood of events around $\e_i$ to build a local descriptor.
The descriptor is then used to test the presence of a corner.
In this work, we consider as in~\cite{Mueggler17} the Time Surface~\cite{Benosman12}
as local descriptor. However, we show how the variance of the standard 
Time Surface formulation is not suited for stable corner detections.
We therefore introduce a normalization scheme to make the time surface invariant to speed.
Moreover, we adopt a machine learning approach and use the invariant time surface 
as input to a classifier trained to discriminate corner events.

\subsection {Speed Invariant Time Surface}
\label{subsec:time_surface}

A common  representation used  in \eb  vision is  the \textit{Surface  of Active
	Events}~\cite{Benosman12},       also      referred       as      \textit{Time
	Surface}~\cite{Lagorce17}.   The  Time Surface  $T$  at  a pixel  $(x,y)$  and
polarity $p$ is defined as
\begin{equation}
	T(x,y,p) \leftarrow t \> ,
	\label{eq:time_surface}
\end{equation}
where $t$  is the time  of the  last event with  polarity $p$ occurred  at pixel
$(x,y)$.  The  Time Surface, besides being  very efficient to compute,  has been
proved       to      be       highly       discriminative      for       several
tasks~\cite{Mueggler17,Sironi18,Zhu18}.  However, we  noticed that local patches
of the Time  Surface can have a  very large variability. In fact,  depending on the
speed, the  direction and the  contrast of the corners,  the aspect of  the time
surface can  vary significantly. 
To  keep the classification model  compact and
efficient, it is thus important to introduce some normalization of its input.

Inspired by~\cite{Mueggler17},  we notice  that only  relative timings,  and not
absolute ones, are  relevant to determine the presence of  a corner. We could therefore
obtain invariance by normalizing 
the times of the events according to their
local speed.  However, normalization  techniques  suitable  for
intensity  images  cannot   be  applied  to  the  time   surface,  as  explained
in~\cite{Alzugaray18b}.  \cite{Alzugaray18b}  proposed to sort the  times of
the events in a  local patch. However, sorting a  local patch at every
incoming  event is  still too  expensive, and  requires to  store multiple  time
stamps per pixel.

Instead, we show how to obtain efficiently a Time Surface that is independent of
the speed,  by keeping a single value for  each pixel location $(x,y)$
and every polarity.  We call this novel formulation \textit{Speed Invariant Time
Surface}.

The intuition for our new formulation goes as follows: Imagine the contour of an
object moving in  one direction. We would  like to capture the  'wave' of events
generated by this  contour, but this wave should have  the same profile whatever
the speed of the contour.  When a  new event arrives, we store a large
value at  its pixel  location in the Time  Surface, and
decrease the values for the surrounding locations. In this way, we progressively
reduce  the   values  for  the   events  previously  generated  by   the  moving
contour. Since  we decrease the  values by a constant  factor, the slope  of the
'wave' is independent of contour's speed.

More precisely,  given a parameter $r$,  we initialize the Speed  Invariant Time
Surface  $S(x,y,p)$ to  $0$ for every  pixel location  $(x,y)$ and
every  polarity $p$. Then,  for every  incoming event
$(x,y,p)$, we consider all pixel locations $(x',y')$ in its neighborhood of size
$(2r+1)\times(2r+1)$.   If  $S(x',y',p)$ is  larger  than  $S(x,y,p)$, then  we
subtract  1 to  $S(x',y',p)$.  Finally,  we set  $S(x,y,p)$ to  $(2r+1)^2$.  The
pseudo-code        of       the        algorithm        is       given        in
Algorithm~\ref{algo:speed_invariant_time_surface}.
\begin{figure}
	\centering
	\includegraphics[width=0.99\columnwidth,height=0.51\columnwidth]{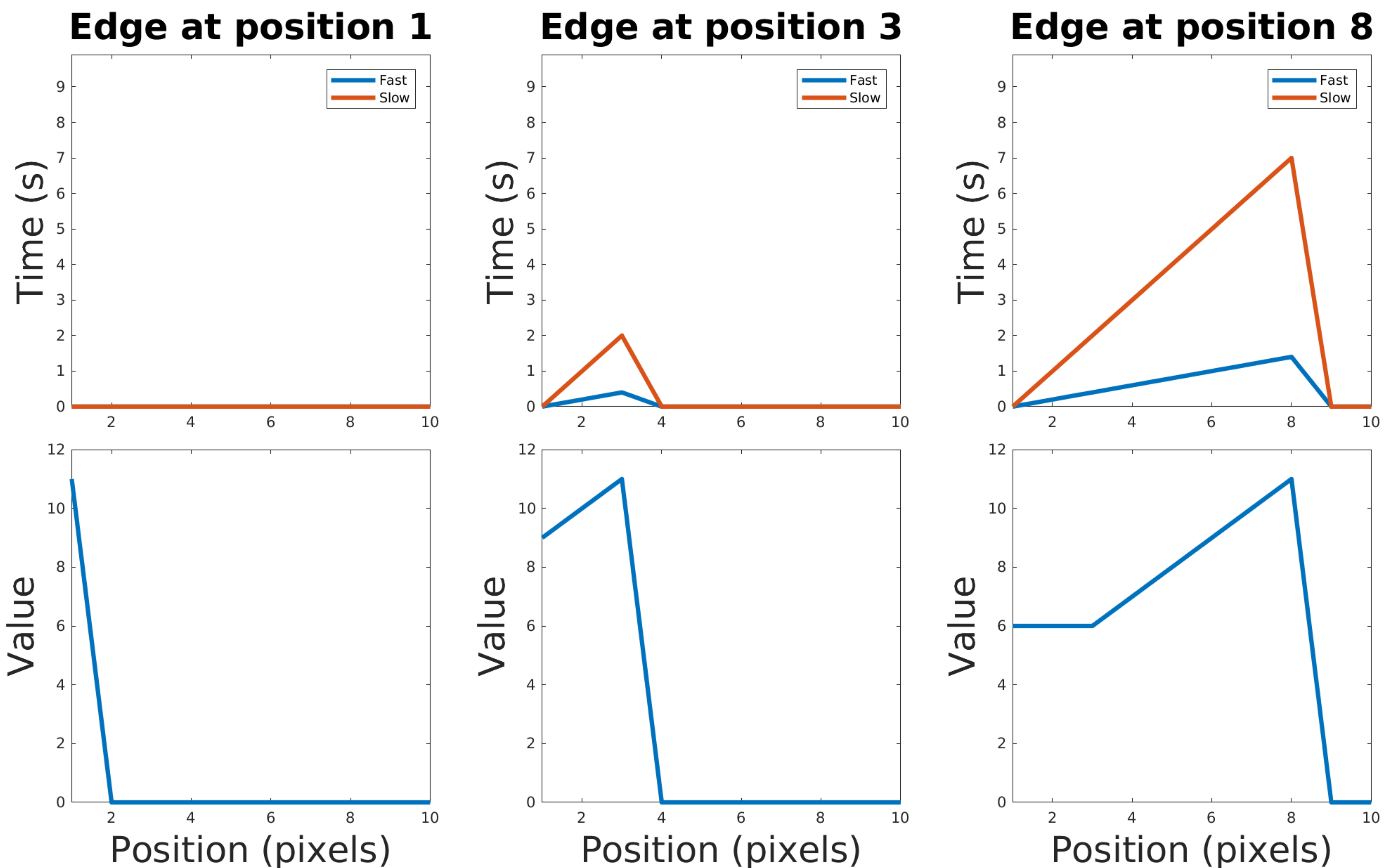}
	\vspace{-2mm}
	\caption{An edge moves from position 1 to 10 at two different speeds. It creates a slope on  
	the Standard Time Surface $T$ {\bf (Top)} and on the corresponding Speed Invariant Time Surface $S$ 
	{\bf (Bottom)}. $S$ is identical for both speeds.}
	\vspace{-5mm}
	\label{fig:comparison_1d}
\end{figure}

 Fig.~\ref{fig:comparison_1d} illustrates  the application  of this  algorithm to
several edges moving from left to right,  each edge moving at a different speed.
For the  standard Time  Surface, each  edge generates  a different  slope behind
it. On  the Speed  Invariant Time  Surface, the slope  is the  same for  all the
edges, and has a  total length of 5 pixels which is equal  to the parameters $r$
used for this experiment. Also, the Speed Invariant Time Surface is constant
after the slope with a value of $r+1$. Finally, it can be seen that the values
in the Speed Invariant Time Surface remain between 0 and $2r+1$.

\begin{figure}
	\centering
	\includegraphics[trim={10cm 4cm 9cm 4cm},clip,width=0.99\columnwidth,height=0.5\columnwidth]{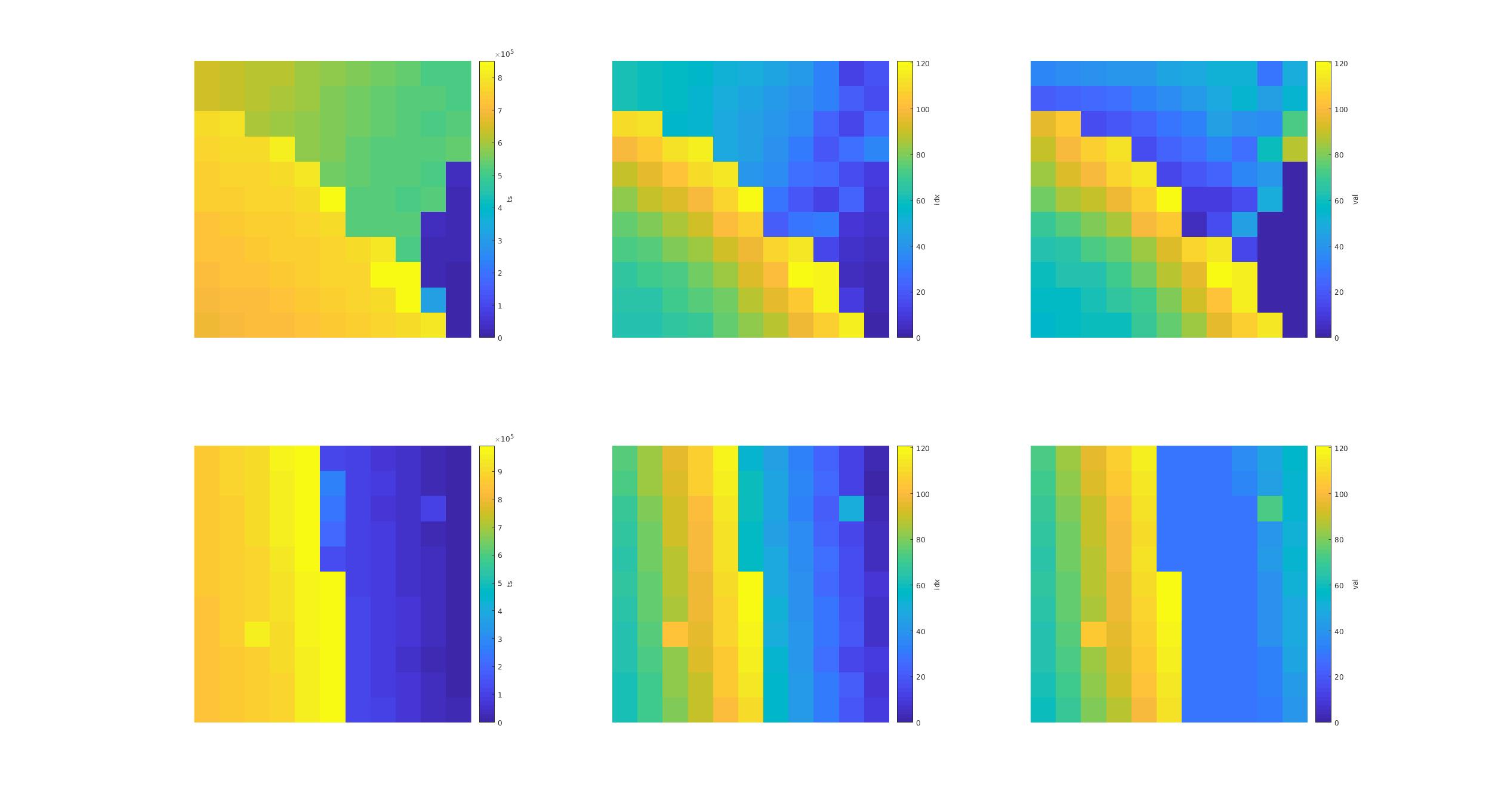}
	\begin{tabular*}{\columnwidth}{c@{\extracolsep{\fill}}c@{\extracolsep{\fill}}c}
			\small{Standard}& \small{Normalization }& \small{Speed Invariant} \\
		\small{Time Surface} &\small{ method of} \cite{Alzugaray18b} & \small{Time Surface} 
	\end{tabular*}
	\vspace{-3mm}
	\caption{Different types of Time Surfaces for two  different patches. The two
		last methods achieve similar results with a high contrast around the edge in
		comparison with the standard Time  Surface. Our Speed Invariant Time Surface
	is much  more efficient to compute and increases the contrast by reducing the values after the edge. }
	\vspace{-4mm}
	\label{fig:comparison_sort_norm}
\end{figure}

In Fig.~\ref{fig:comparison_sort_norm},  we compare  the standard  Time Surface,
the sorting normalization method of~\cite{Alzugaray18b}, and our Speed Invariant
Time Surface.  The two normalization approaches achieve
similar  results for  both examples  with  a high  contrast around  the edge  in
comparison with  the standard  Time Surface.   Furthermore, our  Speed Invariant
Time Surface increases this contrast by reducing the values after the edge.
\vspace{-0mm}
\begin{algorithm}[htbp]				
	\caption{Speed Invariant Time Surface}			
	\label{algo:speed_invariant_time_surface}			
	\begin{algorithmic}[1]
				\State Output: Speed Invariant Time Surface $S(x,y,p)$
		\State Initialization: $S(x,y,p) \leftarrow 0$ for all $(x,y,p)$
		\State For each incoming event $(x,y,p)$, update $S$:
		\For{$-r \leq dx \leq r$}
		\For{$-r \leq dy \leq r$}
		\If{\small{$S(x+dx,y+dy,p) \geq S(x,y,p)$} }
		\State {\small{$S(x+dx,y+dy,p) \leftarrow S(x+dx,y+dy,p) - 1$} }
		\EndIf\EndFor
		\EndFor\State $S(x,y,p) \leftarrow (2r+1)^2$
	\end{algorithmic}
	\vspace{-1mm}
\end{algorithm}

\vspace{-8mm}
\subsection{Learning to Detect Corner Points from Events}
\label{subsec:learnign}
 
Previous corner detectors rely on hand-crafted rules which make them unstable
when the input events do not follow their assumptions.
For example, \evFast is not able to detect corners during a change of motion. 
In the case of \Arc, the detection of obtuse angles lead to a high number of false
detection on edges. Finally, \evHarris is less robust on
fast motions and computationally too expensive. This is why we
train a classifier to detect  corner points from events.  More exactly, given
an incoming event  $\e_i = (x_i,y_i,p_i,t_i)$, we first update  our time surface
$S$ as  explained in Section~\ref{subsec:time_surface},  and we extract from  $S$ a
local patch $\s_i$ of size $n\times n$  centered on the event 2D location $(x_i,
y_i)$.

An ideal detector $\F^*$ taking $\s_i$ as input is defined as
\begin{equation}
	\F^*(\s_i) = \left\{
	\begin{array}{ll}
		1 & \quad \text{if } \e_i \text{ is a feature event, and} \\
		0 & \quad \text{otherwise .}                              
	\end{array} \right.
	\label{eq:classifier}
\end{equation}
\vspace{-2mm}

In principle, any classifier could be used to implement $\F$, but in practice we
chose  a  Random Forest~\cite{Breiman01}  because  of  its efficiency.
For the sake of completeness, in the following we briefly describe Random Forests.
A Random Forest classifier $\F$ is given by an ensemble of decision trees $F_l$:
\begin{equation}
	\F(\s) = \frac{1}{L}\sum_{l=1}^{L}{F_l(\s)} \> .
	\label{eq:trees}
	\vspace{-1mm}
\end{equation}
A  decision tree  $F_l(\s)$ classifies  a  sample $\s  = (s_1,\ldots,s_K)$  by
recursively branching  the nodes of the  tree until the sample  reaches a leaf
node.  In each node a binary split  function is applied to the input to decide
if the sample is sent to the left or the right child node.
  
In  practice, we  use decision  stumps as  split functions~\cite{Criminisi12},
where a descriptor  dimension $d_k$ is compared to a  threshold $th$.  This is
computationally efficient  and effective in  practice~\cite{Criminisi12}.  The
output of the tree is the prediction stored at the leaf reached by the sample,
which in our case is the probability of the sample being a feature point.
  
During training,  each tree is trained  independently in a greedy  manner, one
node at the time.  For a given node, let $\D_N$ be the training set of samples
that reached this node. The optimal $\theta = (k,th)$ for the split
are  obtained  by  minimizing  the Gini  impurity  index~\cite{Friedman01} by exhaustive  search. 
The dataset is then split
in left and right $\D_L$, $\D_R$ and the children nodes are trained recursively.
Training stops when a  maximum depth is
reached, when  a minimum number of  samples is reached or  if the
impurity index is too low.
  
A Random  Forest improves the accuracy  of a single decision  tree by training
multiple  trees. The  trees  need to  be  uncorrelated, and  this  is done  by
randomly sampling  a subset of  the training set for  each tree and  by randomly
subsampling a subset  of the descriptors dimensions at each  node.

\comment{
$\mathcal{N}(\e_i) = \{\e_j : t_j \leq t_i\}$ is the set of events used as input to 
to the classification method, which will be used in our case to build 
the Speed Invariant Time Surface descriptor,
and the binary  class $c_i$  of event
$\e_i$  is 1  if $\e_i$  is a  feature event,  0 otherwise.\vincentrmk{there  is
	probably regions of the time surface around the point somewhere}\amosrmk{yes, in theory
	the whole stream of events up to $\e_i$ can be used. Modified above.}.
}

\vspace{-1mm}

\subsection{Building a Dataset for \Eb Corner Detection}
\label{subsec:train_dataset}

To train our classifier $\F$, we  need a labeled training set $\{(\s_j, c_j)\}$,
where the binary class $c_j$ of event $\e_j$  is 1 if $\e_j$ is a feature event,
0 otherwise.  Building such a dataset in the case of \eb cameras is not trivial.
Manual annotation is impractical, since it  would require to label single events
in a sequence of million of events per second.

We propose to leverage the graylevel measurement provided by a HVGA ATIS sensor,
already discussed in Section~\ref{sec:event-based_cameras}.  For every event, we
apply the Harris corner detector to its location only, in the graylevel image.
If a corner is detected  at this location, we add the event to our training set
as a positive sample, and as a negative sample otherwise.
In practice, even the Harris detector  applied to graylevel images can sometimes
fail in presence  of noise, and we  therefore decided to acquire  our dataset by
recording    well     contrasted    geometric     patterns. 
Examples of this dataset are shown in the supplemen
tary material
The dataset can be downloaded at the following URL\footnote{ \texttt{http://prophesee.ai/hvga-atis-corner-dataset}}.
In  this  way,  we greatly  reduce  the  number  of
outliers in the dataset.  As we will see in Section~\ref{sec:experiments}, even if the
pattern  we  use  is quite  simple,  the  classifier  learnt  on such  data  can
generalize well in more complex real situations. \amos{We also extend this 
	protocol to a DAVIS sensor. In this case the corners are detected on entire frames
	at regular intervals and all the events falling within 2 pixels form a detected corner 
	in a time interval of $5ms$
are marked as corner events.}

\section{Evaluation of Event-based Detectors}
\label{sec:datasets}

\paragraph{Event-based Datasets for Corner Detection}
\label{subsec:test_datasets}

We evaluate  our method  on two \eb  datasets.  The first  is the  commonly used
Event-Camera  dataset~\cite{Mueggler17b}.   It  is  composed  by  recordings  of
different scenes  taken with a  DAVIS sensor. As done  by previous  works, 
we consider  the
\texttt{boxes, shapes, hdr} and \texttt{dynamic} subsets for evaluation,
for which camera  poses groundtruth are provided, 
\amos{but we keep the simplest \texttt{shapes} dataset to train our model.}

The second dataset we use is a new dataset that we introduce for the first time in this paper. 
Compared to the Event-Camera dataset, ours was acquired using a higher resolution 
HVGA ATIS sensor, 
thus allowing better localization and finer feature detection,
together with higher data rate. 
While the Event-Camera dataset is more generic and adapted to test visual
odometry and SLAM applications, 
the intent of our dataset is to specifically evaluate the accuracy of 
\eb feature detection and tracking algorithms.
We call this dataset the \textit{HVGA ATIS Corner Dataset}.
It consists of 7 sequences  of increasing difficulty, from  a standard
checkerboard to a complex natural image (Fig.~\ref{fig:comparison_detections}).
We record planar
patterns, so that ground truth acquisition  is simple and the evaluation is less
affected by triangulation errors.
\vspace{-4mm}
\paragraph{Evaluation Metrics}
\label{subsec:metrics}

Previous approaches for  \eb feature detection rely on  Harris corners extracted
from  graylevel  images~\cite{Vasco16,Zhu17,Gehrig18}.   This approach  has  the
shortcoming  of  penalizing \eb  features  when  not  matched to  any  graylevel
one, even  if these features might  correspond to stable  points in
the events stream. We notice that our  approach, since it  was trained starting from  Harris corners
would have an advantage when using this metric.

Other  works~\cite{Mueggler17} instead, consider a detected corner valid if, 
when associated with nearby corners, it forms a well localized track in space and
time.  This metric evaluates how easy it  is to track a given feature.  However,
it does not take into account the correct trajectory of the feature.

We also start from the observation that a stable detector, able to continuously
identify a  feature in the  events stream, would remove 
the  need  of  complex  tracking  and  data
association methods. 
Therefore, in order to assess also the quality of  an \eb detector, we combine it with
a simple  tracking algorithm  based on  nearest neighbor  matching in  space and
time.
After  tracking,  we can evaluate  the  accuracy  of  the  method by  computing  the
reprojection error associated to the feature tracks. 
	
In the case of the planar dataset,
the  reprojection error  is  computed  by estimating  a  homography between  two
different timestamps.  More precisely, given  two timestamps $t_1$ and $t_2$, we
collect all the  features that fall within a $5ms$  timewindow from these times.
We then consider the last feature for each track. This gives two sets of 2D
correspondences that  can be
used to  estimate the homography  between the two views  of the camera.   We use
RANSAC for a  robust estimation.  Once the homography is  computed, we reproject
points from time $t_2$ to the reference  time  $t_1$ and compute the  average
distance between  the reference points and the projected ones. During this process, 
we exclude points detected outside the planar pattern.

In the case of the Event-Camera dataset, which contains 3D scenes, the reprojection error 
is computed by triangulating the tracked points, using the available 3d poses. 
We use the same protocol as in~\cite{Alzugaray18b} and report also the percentage
of tracks with an error smaller than 5 pixels.
Finally we compare the methods in terms of computational
time, as done in~\cite{Gehrig18, Alzugaray18b}. 
All our experiments were implemented in C++ and run on a Xeon CPU E5-2603 v4 at 1.70GHz.
 
\section{Experiments}
\label{sec:experiments}

\amos{
	\paragraph{Parameters and Baselines}
																				
	Our method depends on few parameters, namely the radius $r$  used to compute the 
	Speed Invariant Time Surface, the size $n$ of the classifier input patch, 
	and the parameters for the Random Forest.
		The parameters were optimized by cross-validation on the training set of
	Section~\ref{subsec:train_dataset}  to  minimize  the  misclassification
	error.  Once  the best parameters have  been found, we fix  them and use
	them for all the test sequences of Section~\ref{subsec:test_datasets}.
		For  the descriptor, we  set $r=6$  and $n  = 8$.   For the
	Random Forest, we use 10 trees. We stop growing the trees when there are
	less than 50 training samples in a branch.

	We  compare   our  method  against  previously   published  \eb  feature
	detectors: the  \eb Harris  detector of~\cite{Vasco16}, which  we denote
	\evHarris; the  \eb Fast of~\cite{Mueggler17} \evFast;  and its recent modification
	\Arc~\cite{Alzugaray18}.   For all of  these methods we  use the
	publicly available implementation provided by the authors.
																				
	As  done in~\cite{Alzugaray18},  we  also apply  an  \eb 'trail'  filter
	before each detector.  This filter removes the multiple events generated
	by  a  contrast  step  in  intensity  which  caused  multiple  threshold
	crossing.    The  timings   of  Table~\ref{tab:computational_time}   are
	reported taking into account this filtering.
				
	\begin{figure*}
	\centering
  \includegraphics[width=0.99\linewidth]{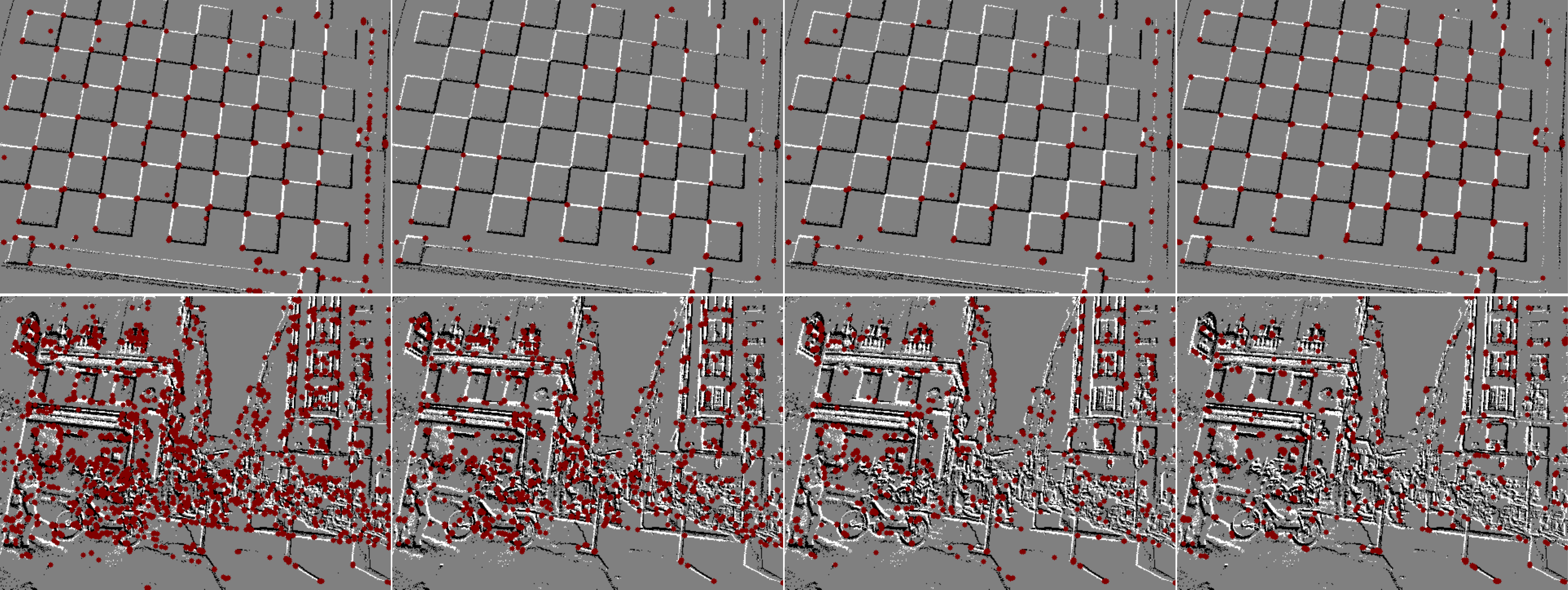}
  \begin{tabular*}{\columnwidth}{cccc}
		\hspace{-3cm}\Arc~\cite{Alzugaray18} &\hspace{2.1cm} \evHarris~\cite{Vasco16} &\hspace{2.2cm} \evFast~\cite{Mueggler17} &\hspace{2.3cm} SILC
  \end{tabular*}
  \vspace{-3mm}
	\caption{Comparison of \eb detectors on the HVGA ATIS Corner Dataset. 
  Positive and negative events accumulated during 5ms showed in black and white with 
  corresponding corner events in red. Our method is more robust to noise and can detect stable corners even in complex scenes.}
  \vspace{-4mm}
	\label{fig:comparison_detections}
\end{figure*}

 	\vspace{-3mm}						
	\paragraph{Ablation Study}
	In the first set of experiments,  we quantify the improvement brought by
	our Speed Invariant Time  Surface formulation against  the standard
	Time Surface.
										
	We train two Random Forests, one  using local patches extracted from the
	Time Surface of~\cite{Benosman14}, and  the second  one on
	Speed Invariant Time Surface patches.  Then, we apply these detectors to
	the HVGA dataset and track the features using nearest
	neighbor  data association  with a  radius of  3 pixels  and a  temporal
	window of $10ms$.  From the tracked features we  estimate the homography
	using different time steps, as explained in Section~\ref{subsec:metrics}.
	The    corresponding    reprojection     errors    are    reported    in
	Table~\ref{tab:ablation_study}.   As we  can  see our  method has  lower
	reprojection error, the reason is that  the detector trained on the Time
	Surface can not  generalize well, and, a part for  the simple chessboard
	pattern returns  very noisy  detections. We  refer to  the supplementary
	material for a visual comparison of the detected features.
{\small
	\begin{table}[htpb]
		\caption{Reprojection error  on the  HVGA Corner dataset  for different
			values of $\Delta  t$ used  to estimate  the homography,
			when    training    a    Random    Forest    on    the    Time
			Surface~\cite{Benosman14} or the proposed Speed Invariant Time
			Surface (SILC).   SILC results in  a more stable  and accurate
		detections.}
		\vspace{-5mm}
		\begin{center}
			\tabcolsep=0.11cm
			\begin{tabular}{@{}l cccc@{}}
				\toprule
				{Random Forest}                & $\Delta t = 25ms$ & $\Delta t = 50ms$ & $\Delta t = 100ms$ \\
				\hline
				{Using $T$~\cite{Benosman14} } & 5.79              & 8.48              & 12.26              \\
				{Using $S$ (SILC)}             & {\bf 2.45}        & {\bf 3.03 }       & {\bf 3.70  }       \\
				\bottomrule
			\end{tabular}		\end{center}
		\label{tab:ablation_study}
		\vspace{-0mm}
	\end{table}
	}
	\vspace{-9mm}	
	\paragraph{Evaluation on the HVGA ATIS Corner Dataset}
	\begin{figure}
	\centering
	\begin{tabular}{@{\hspace{0cm}}c@{\hspace{0cm}}}		\includegraphics[width=1\linewidth]{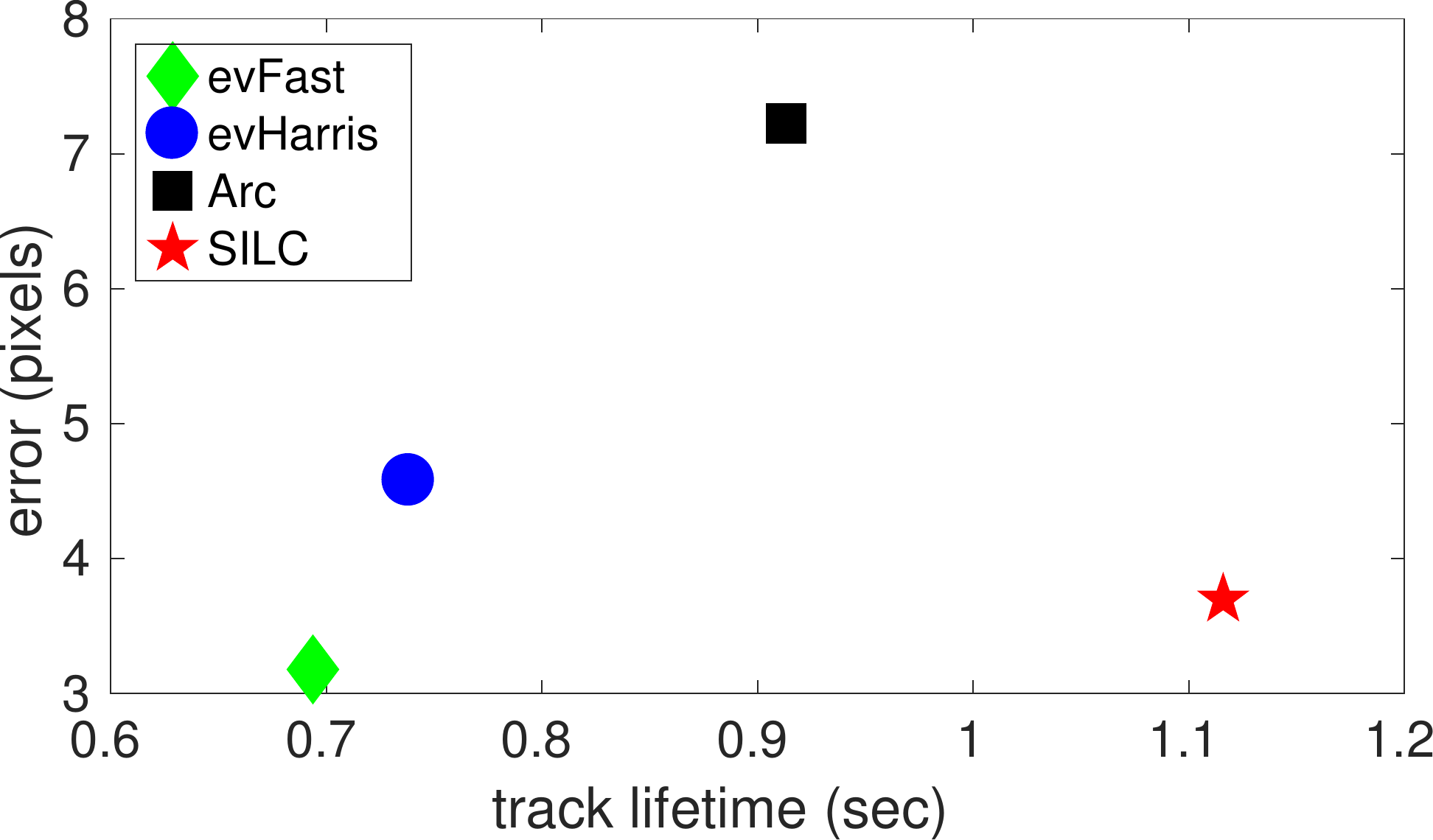} 
	\end{tabular}
	\vspace{-4mm}
	\caption{Comparison of detector performance on the HVGA ATIS Corner dataset.
		Our method is able to track a corner longer than the 
  baselines while keeping a low reprojection error.
  }
	\vspace{-7mm}
	\label{fig:error_vs_lifetime_planar}
\end{figure}
     							
	In  this section,  we compare  the  stability of  \eb feature  detectors
	against the  baseline methods  using the homography  reprojection error.
	Since this metric  can be applied only  to planar scenes, we  use it for
	our dataset and  not on the Event-Camera dataset.   For nearest neighbor
	data association, we used a radius of  3 pixels and a temporal window of
	$10ms$.
					We also  compute the lifetime  of the  tracked features, defined  as the
	difference between  the first and the  last detection of the  track.  We
	then compute the average lifetime of the first 100 features.
										
	We  compute the  homography at  different time  intervals, ranging  from
	$25ms$   to   $200ms$.   The   results  for   $100ms$   are   shown   in
	Fig.~\ref{fig:error_vs_lifetime_planar}.   
	Detailed values  for other
	values and for each sequence are reported in the supplementary material,
	and show similar  behavior.  
	Our method is able to  track a corner point
	longer while having a low reprojection  error.  The error for \evFast is
	slightly  lower,  but  the  tracks for  this  method  are  significantly
	shorter.  Finally,  \evHarris and \Arc  are very sensitive  to noise
	and respond on  edges, which make their tracks longer  then \evFast, but
	poorly reliable.
										
	Qualitative results are shown in Fig.~\ref{fig:error_vs_lifetime_planar}. 
	A snapshot example can be found in Fig.~\ref{fig:comparison_detections}.
	Corresponding video sequences are provided in the supplementary material.
							
	\vspace{-5mm}
	
	\paragraph{Evaluation on the Event-Camera Dataset}
															   
	We repeat similar  experiments for the Event-Camera  Dataset. Since the
	resolution of  the camera  is lower  for this dataset,  we use  a larger
	temporal window, equal to 50ms, for  the tracking.  Because of the lower
	resolution  and  the  higher  level  of  noise,  this  dataset  is  more
	challenging.

	The reprojection  error and the percentage of valid tracks
	are   given   in  Tab.~\ref{tab:eb_dataset}.
	Detailed  values are reported  in the  supplementary
	material.						
	We notice that applying the model trained on the ATIS camera (SILC ATIS) generalizes well on the DAVIS, 
	reaching similar performance than the baseline.
	Since the DAVIS camera provides gray-level images, 
	we also retrained a Random Forest on the DAVIS data (SILC DAVIS).
	We use a small model composed of 10 trees of depth 10.
					We observe that the results obtained with our detector and a simple nearest neighbor tracking are 
	comparable with results obtained with complex trackers~\cite{Alzugaray18b}.
							\vspace{-2mm}
	\begin{table}[htpb]
		\caption{Evaluation on the Event-camera dataset.
		 Our method has the lowest reprojection error and generalizes well across different sensors.}
		\vspace{-5mm}
		\begin{center}
			\tabcolsep=0.11cm
			\begin{tabular}{@{}l ccccc@{}}
				\toprule
				\textbf{}                & \evHarris      & \evFast           & \Arc               & SILC   & SILC       \\
				\textbf{}                & \cite{Vasco16} & \cite{Mueggler17} & \cite{Alzugaray18} & {ATIS} & {DAVIS}    \\
				\hline
				{3D reprj. error (pix) } & 2.46           & 2.50              & 2.58               & 2.53   & \bf{ 2.16} \\
				{Valid tracks ($\%$)}    & 47.4           & 50.1              & 42.9               & 47.3   & {\bf 65.3} \\
				\bottomrule
			\end{tabular}		\end{center}
		\label{tab:eb_dataset}		
		\vspace{-6mm}
	\end{table}

																															\begin{table}[htpb]
		\caption{Computational Time on the HVGA dataset.
		 Our method is real time despite the high data rate of the sensor. }
		\vspace{-5mm}
		\begin{center}
			\tabcolsep=0.11cm
			\begin{tabular}{@{}l ccccc@{}}
				\toprule
				\textbf{}              & \evHarris      & \evFast           & \Arc               & SILC   \\
				\textbf{}              & \cite{Vasco16} & \cite{Mueggler17} & \cite{Alzugaray18} & {} \\
				\hline
				{Event rate ($Mev/s$)} & 0.22           & 1.74              & 5.61               & 1.61   \\
								{Real Time Factor}     & 0.60           & 3.80              & 12.32              & 3.53   \\
				\bottomrule
			\end{tabular}		\end{center}
		\label{tab:computational_time}
		\vspace{-10mm}
	\end{table}

}
 
\section{Conclusion and Future Work}
\vspace{-1mm}
\label{sec:conclusion}
We presented an efficient and accurate learning approach for \eb corner detection.
Our method produces more stable and repeatable corners compared to the state of-the-art
and when coupled with a simple tracking algorithm gives good accuracy.
A key component for our approach is a novel formulation of the Time Surface,
which provides a rich \eb representation which is invariant to the local speed of the object.
In the future, we plan to apply the Speed Invariant Time Surface to other
\eb vision tasks, such as low-latency object detection or relocalization.

\begin{acknowledgement}
    Dr. Vincent Lepetit is a senior member of the \emph{Institut Universitaire de France}.
\end{acknowledgement}

{\small
\bibliographystyle{ieee}

}

\onecolumn

\section*{Appendix}

\begin{table}[H]
	\centering
	\begin{tabular}{l|cc}
	
	\hline
	& Gen3 CD           & Gen3 ATIS         \\ \hline
	Supplier                           & Prophesee         & Prophesee         \\
	Year                               & 2017              & 2017              \\
	Resolution (pixels)                & 640x480           & 480x360           \\
	Latency ($\mu s$)                    & 40 - 200          & 40 -200               \\
	Dynamic range (dB)                 & \textgreater{}120 & \textgreater{}120 \\
	Min. contrast sensitivity (\%)     & 12                & 12                \\
	Die power consumption (mW)         & 36 - 95           & 25 - 87           \\
	Camera Max. Bandwidth (Meps)       & 66                & 66                \\
	Chip size ($mm^2$)                    & 9.6x7.2           & 9.6x7.2           \\
	Pixel size ($\mu m^2$)                   & 15x15             & 20x20             \\
	Fill factor (\%)                   & 25                & 25                \\
	Supply voltage (V)                 & 1.8               & 1.8               \\
	Stationary noise (ev=pix=s) at 25C & 0.1               & 0.1               \\
	CMOS technology ($\mu m$)               & 0.18              & 0.18              \\
																			& 1P6M CIS          & 1P6M CIS          \\ \hline
	Grayscale output                   & no                & yes               \\
	Grayscale dynamic range (dB)       & NA                & \textgreater{}100 \\
	Max. framerate (fps)               & NA                & NA                \\ \hline
	IMU output                         & 1 kHz             & 1 kHz             \\ \hline
	\end{tabular}
	\begin{center}
		\caption{Technical description of the cameras used for the paper. The Gen3 ATIS was used to generate the ground truth.}
	\end{center}

\end{table}	

\end{document}